\documentclass[10pt,twocolumn,letterpaper]{article}
\pdfoutput=1
\makeindex

\usepackage{iccv}
\usepackage{times}
\usepackage{epsfig}
\usepackage{graphicx}
\usepackage{amsmath}
\usepackage{amssymb}
\usepackage{epstopdf}
\usepackage{dsfont}
\usepackage{color}

\usepackage[pagebackref=true,breaklinks=true,letterpaper=true,colorlinks,bookmarks=false]{hyperref}
\bgroup\catcode`\#=12\egroup  

\def\iccvPaperID{11} 

\iccvfinalcopy
\ificcvfinal\fi
\usepackage{amsthm}
\newtheorem{property}{Property}
\newtheorem{definition}{Definition}

\usepackage[pagebackref=true,breaklinks=true,letterpaper=true,colorlinks,bookmarks=false]{hyperref}
\usepackage{amsthm}

\def \bbR{{\mathbb{R}}}
\def \cS{{\mathcal{S}}}
\def \cD{{\mathcal{D}}}

\DeclareMathOperator*{\argmax}{arg\,max}

\begin{document}
	
	\title{A Strategy for an Uncompromising Incremental Learner}
	
	\author{Ragav Venkatesan, Hemanth Venkateswara, Sethuraman Panchanathan, and Baoxin Li
		\\
		Arizona State University, Tempe, AZ, USA \\
		{\tt\small ragav.venkatesan@asu.edu, hemanthv@asu.edu, panch@asu.edu, baoxin.li@asu.edu}}
	
	\maketitle

	\begin{abstract}
		Multi-class supervised learning systems require the knowledge of the entire range of labels they predict. Often when learnt incrementally, they suffer from catastrophic forgetting. To avoid this, generous leeways have to be made to the philosophy of incremental learning that either forces a part of the machine to not learn, or to retrain the machine again with a selection of the historic data. While these hacks work to various degrees, they do not adhere to the spirit of incremental learning. In this article, we redefine incremental learning with stringent conditions that do not allow for any undesirable relaxations and assumptions. We design a strategy involving generative models and the distillation of dark knowledge as a means of hallucinating data along with appropriate targets from past distributions. We call this technique, \emph{phantom sampling}.We show that phantom sampling helps avoid catastrophic forgetting during incremental learning. Using an implementation based on deep neural networks, we demonstrate that phantom sampling dramatically avoids catastrophic forgetting. We apply these strategies to competitive multi-class incremental learning of deep neural networks. Using various benchmark datasets and through our strategy, we demonstrate that strict incremental learning could be achieved. We further put our strategy to test on challenging cases, including cross-domain increments and incrementing on a novel label space. We also propose a trivial extension to unbounded-continual learning and identify potential for future development.
		
	\end{abstract}

	\section{Introduction}
	\label{sec:intro}
	
	Animals and humans learn incrementally. 
	A child grows its vocabulary of identifiable concepts as different concepts are presented, without forgetting the concepts with which they are already familiar. 
	Antithetically, most supervised learning systems work under the omniscience of the existence of all classes to be learned, prior to training.
	This is crucial for learning systems that produce an inference as a conditional probability distribution over all known categories.
	
	Incremental supervised learning though reasonably studied, lacks a formal and structured definition. 
	One of the earliest formalization of incremental learning comes from the work of Jantke~\cite{jantke1993types}. 
	In this article the author defines incremental learning roughly as systems that \emph{``have no permission to look back at the whole history of information presented during the learning process''}. 
	Immediately following this statement though is the relaxation of the definition: \emph{``Operationally incremental learning algorithms may have permission to look back, but they are not allowed to use information of the past in some effective way''}, with the terms \emph{information} and \emph{effective} not being sufficiently well-defined.  
	Subsequently, other studies made conforming or divergent assumptions and relaxations thereby adopting their own characteristic definitions.
	Following suit, we redefine a more fundamental and rigorous incremental learning system using two fundamental philosophies: data membrane and domain agnosticism. 
	
	\begin{figure*}[!ht]
		\begin{center}
			\includegraphics[width=0.99\linewidth]{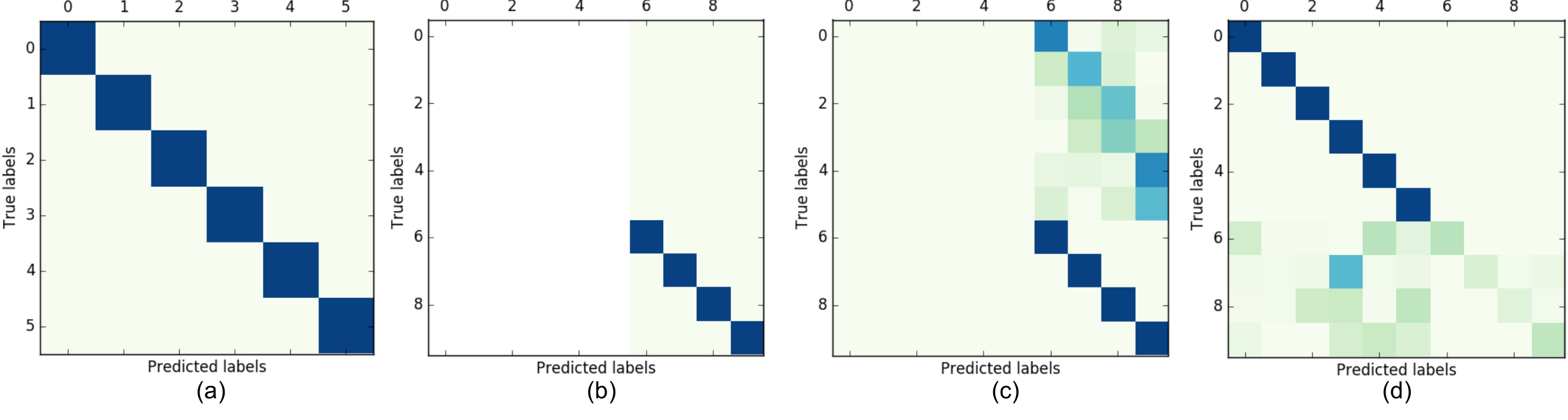}
		\end{center}
		
		\caption{Catastrophic forgetting: Figure (a) is the confusion matrix of a network $N_b$, trained and tested on data from a subset containing only samples of labels $0 \hdots 5$. 
			Figure (b) is the confusion matrix of a network initialized with the weights of trained $N_b$, re-trained with data from classes $6 \hdots 9$ and tested on the same label space. No testing samples were provided for the classes $0 \hdots 5$. 
			Figure (c) is the same network as (b) tested on the entire label space. Figure (d) is similar to (c) but trained with a much lower learning rate. These confusion matrices demonstrate that a neural network retrained on new labels without supplying it data from the old data subset, forgets the previous data, unless the learning rate is very measured and slow as was the case in (d). 
			If the learning rate were slow, though the old labels are not forgotten, new labels are not effectively learned.}
		
		\label{fig:forgetting}
	\end{figure*}
	
	Consider there are two sites: the base site $\cS_b$ and the incremental site $\cS_i$ each with ample computational resources.
	$\cS_b$ possesses the base dataset $\cD_b = \{(x_l^b,y_l^b), l\in \{1,2, \hdots n\}\}$, where $x_l^b \in {\bbR^d}, \forall l$ and $y_l^b \in \{1,2, \hdots j\}, \forall l$.
	$\cS_i$ possesses the increment dataset $\cD_i=\{(x_l^i,y_l^i), l\in \{1,2, \hdots m\}\}$, where $x_l^i \in {\bbR^d}, \forall l$ and $y_l^i \in \{j+1, j+2, \hdots c\}, \forall l$ and $y_l^i \not\in \{0, 1, \hdots j\}, \forall l$.
	
	\begin{property}
		\label{prop:membrane}
		$\cD_b$ is only available at $\cS_b$ and  $\cD_i$ is only available at $\cS_i$. 
		Neither set can be transferred either directly or as features extracted by any deterministic encoder, either in whole or in part to the other site, respectively. 
	\end{property}
	$\cS_b$ is allowed to trai n a discriminative learner $N_b$ using $\cD_b$ and make $N_b$ available to the world. 
	Once broadcast, $\cS_b$ does not maintain $N_b$ and will therefore not support queries regarding $N_b$. Property~\ref{prop:membrane} is referred to as the \emph{data membrane}. 
	Data membrane ensures that $\cS_i$ does not query $\cS_b$ and that no data is transferred either in original form or in any encoded fashion (say as feature vectors). 
	The generalization set at $\cS_i$ contains labels in the space of  $y \in\{1 \hdots c\}$. 
	This implies that though $\cS_i$, has no data for training the labels $1 \hdots j$, the discriminator $N_i$ trained at $\cS_i$ with $\cD_i$ alone is expected to generalize on the combined label space in the range $1 \hdots c$. 
	$\cS_i$ can acquire $N_b$ and other models from $\cS_b$ and infer the existence of the classes $y \in \{1,2, \hdots j\}$ that $N_b$ can distinguish. Therefore incremental learning differs from the problem of zero-shot novel class identification.
	
	A second property of multi-class incremental learning is domain agnosticism, which can be defined as follows:	
	\begin{property}
		\label{prop:agnostic}
		No priors shall be established as to the dependencies of classes or domains between $\cD_b$ and $\cD_i$. 
	\end{property}
	Property~\ref{prop:agnostic} implies that we cannot presume to gain any knowledge about the label space of $\cD_b$ ($\{0 \hdots j\}$) by simply studying the behaviour of $N_b$ using $\cD_i$.
	In other words, the predictions of the network $N_b$ does not provide us meaningful enough information regarding $\cD_i$. 
	This implies that the conditional probability distribution across the labels in $y \in \{0 \hdots j\}$, $P_{N_b}(y|x)$ for $(x,y)\in\cD_i$ produced by $N_b$, cannot provide any meaningful inference to the conditional probability distribution across the labels $y \in \{j+1 \hdots c\}$ when generalizing on the incremental data. 
	For any samples $x \in \cD_i$, the conditional probability over the labels of classes $y \in \{0 \hdots j\}$ are meaningless. 
	Property (\ref{prop:agnostic}) is called \emph{domain agnosticism}. 
	
	From the above definition it is implied that sites must train independently. 
	The training at $\cS_i$ of labels $y \in \{j+1 \hdots c\}$ could be at any state when $\cS_b$ triggers site $\cS_i$ by publishing its models, which marks the beginning of incremental training at $\cS_i$. 
	To keep experiments and discussions simpler, we assume the worst case scenario where the site $2$ does not begin training by itself, but we will generalize to all chronology in the later sections.
	
	We live in a world of data abundance. 
	Even in this environment of data affluence, we may still encounter cases of scarcity of data. 
	Data is a valuable commodity and is often jealously guarded by those who posses it. 
	Most large institutions and organizations that deploy trained models, do not share the data with which the models are trained.
	A consumer who wants to add additional capability is faced with an incremental learning problem as defined. 
	In other cases, such as in military or medicine, data may be protected by legal, intellectual property and privacy restrictions. 
	A medical facility that wants to add the capability of diagnosing a related-but-different pathology to an already purchased model also faces a similar problem and often has to expend large sums of money to purchase an instrument with this incrementally additional capability. 
	All these scenarios are plausible contenders for strict incremental learning following the above definition.
	The data membrane property ensures that even if data could be transferred, we are restricted by means other than technological, be it legal or privacy-related that prevents the sharing of data across sites. 
	The domain agnosticism property implies that we should be able to add the capability of predicting labels to the network, without making any assumptions that the new labels may or may not hold any tangible relationship to the old labels. 
	
	\noindent\textbf{A trivial baseline}: Given this formalism, the most trivial incremental training protocol would be to train a machine at $\cS_b$ with $\cD_b$, transfer this machine (make it available in some fashion) to $\cS_i$. 
	At $\cS_i$, initialize a new machine with the parameters of the transferred machine, while alerting the new machine to the existence of classes $j+1, \hdots c$ and simply teach it to model an updated conditional probability distribution over classes $\{1,2, \hdots c\}$. 
	A quick experiment can demonstrate to us that such a system is afflicted by a well-studied problem called catastrophic forgetting. 
	Figure~\ref{fig:forgetting} demonstrates this effect using neural networks. 
	This demonstrates that without supplying samples from $\cD_b$, incremental training without catastrophic forgetting at $\cS_i$ is difficult without relaxing our definition. 
	
	To avoid this, we propose that the use of generative models trained at $\cS_b$, be deployed at $\cS_i$ to hallucinate samples from $\cD_b$. 
	The one-time broadcast from $\cS_b$ could include this generator along with the initializer machine that is transferred. 
	While this system could generate samples-on-demand, we still do not have targets for the generated samples to learn classification with. 
	To solve this problem, we propose the generation of supervision from the initializer network itself using a temperature-raised softmax.
	A temperature raised softmax was previously proposed as a means of distilling knowledge in the context of neural network compression~\cite{hinton2015distilling}. 
	Not only does this provide supervision for generated samples, but will also serve as a regularizer while training a machine at $\cS_i$, similar to the fashion described in~\cite{hinton2015distilling}.
	
	In summary this paper provides two major contributions: 1. A novel, uncompromising and practical definition of incremental learning and 2. a strategy to attack the defined paradigm through a novel sampling process called \emph{phantom sampling}. The rest of this article is organized as follows: section~\ref{sec:proposed} outlines the proposed method, section~\ref{sec:lit} discusses related works on the basis of the properties we have presented, section~\ref{sec:res} presents the design of our experiments along with the results, section~\ref{sec:continual} extends this idea to continual learning systems, where we present an trivial extension to more than one increment and section~\ref{sec:conclusion} provides concluding remarks.

	\section{Proposed method}
	\label{sec:proposed}
	
	\begin{figure*}[!]
		\begin{center}
			\includegraphics[width=0.999\linewidth]{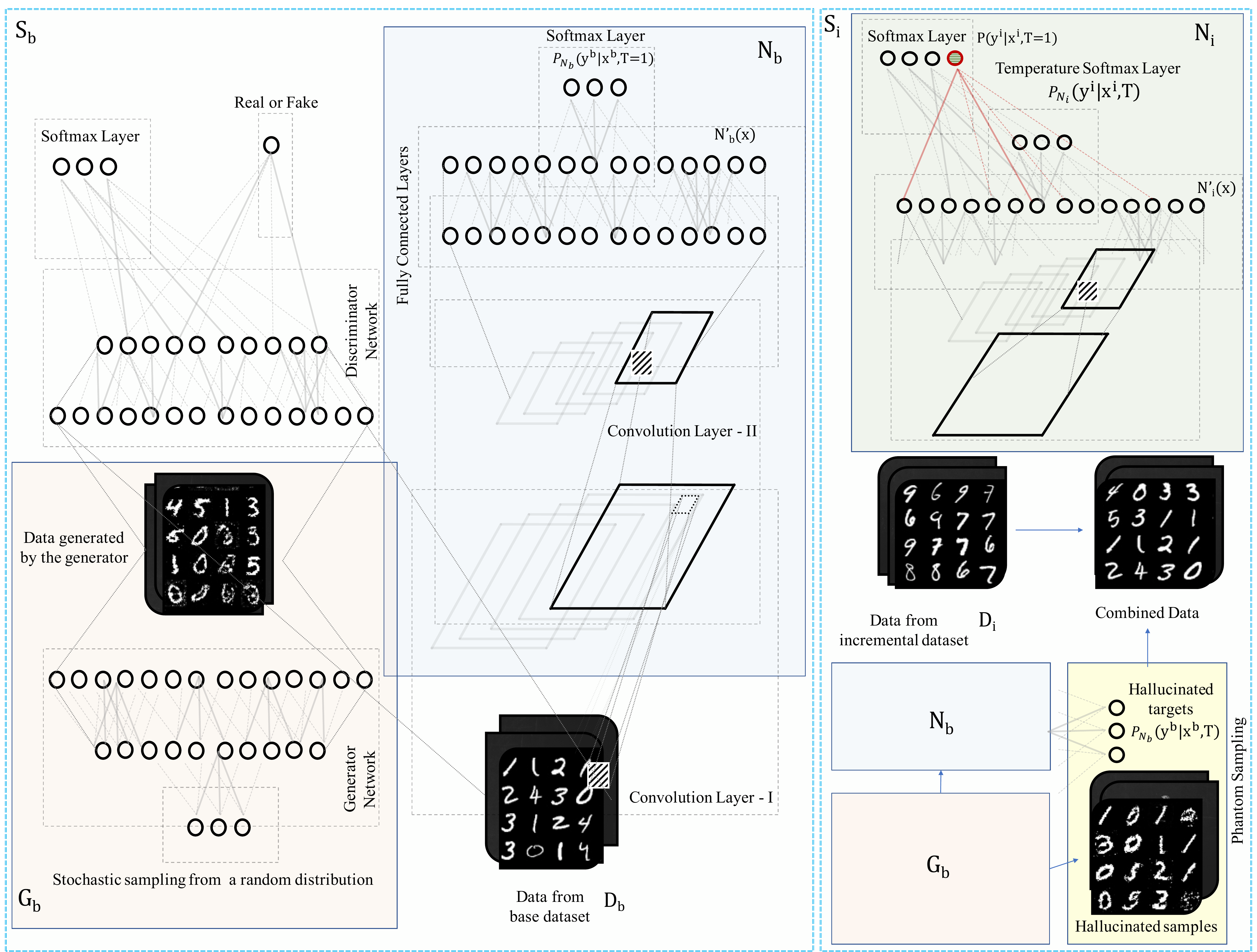}
		\end{center}
		\caption{Sites $\cS_b$, $\cS_i$ and the networks that they train respectively. The networks $G_b$ and $N_b$ are transferred from $\cS_b$ to $\cS_i$ and work in feed-forward mode only at $\cS_i$. In this illustration using MNIST dataset, $j=5$. Classes $[0\hdots5]$ are in $\cD_b$ and classes $[6 \hdots 9]$ are available in $\cD_i$}
		\label{fig:proposed-architecture}
	\end{figure*}
	
	Our design begins at $\cS_b$. 
	Although $\cS_b$ and $\cS_i$ may train at various speeds and begin at various times, in this presentation we focus on the systems that mimic the following chronology of events:
	\begin{enumerate}
		\item $\cS_b$ trains a generative model $G_b$ and a discriminative model $N_b$  for $P(x^b)$ and $P_{N_b}(y \vert x^b)$ using $(x^b,y^b) \in \cD_b$, respectively.
		\item $\cS_b$ broadcasts $G_b$ and $N_b$.
		\item $\cS_i$ collects the models $G_b$ and $N_b$ and initializes new model $N_i$ with the parameters of $N_b$ adding new random parameters as appropriate. 
		Expansion using new random parameters is required since, $N_i$ should make predictions on a larger range of labels.
		\item Using $\cD_i$ together with phantom sampling from $G_b$ and $N_b$, $\cS_i$ trains the model $N_i$ until convergence.
	\end{enumerate} 
	This is an asymptotic special case of the definition established in the previous section and is therefore considered. 
	Other designs could also be established and we will describe briefly a generalized approach in the latter part of this section. 
	While the strategy we propose could be generalized to any discriminatory multi-class classifier, for the sake of clarity and being precise, in this article we restrict our discussions to the context of deep neural networks. 
	
	The generative model, $G_b$ models $P(x | \cD_b)$. 
	In this article we considered networks that are trained as simple generative adversarial networks (GAN) for our generative models. 
	GANs have recently become very popular for approximating and sampling from distributions of data. 
	GAN was originally proposed by Goodfellow et. al, in 2014 and has since seen many advances~\cite{goodfellow2014generative}.  
	We consider the GANs proposed in the original article by Goodfellow et. al, for the sake of convenience.
	We use a simple convolutional neural network model as the discriminator $N_b$.  
	Figure~\ref{fig:proposed-architecture} shows the overall architecture of our strategy with $G_b$ and $N_b$ within the $\cS_b$ capsule. 
	As can be seen, $G_b$ attempts to produce samples that are similar to the data and $N_b$ learns a classifier using the softmax layer that is capable of producing $P_{N_b}(y^b \vert x^b)$ as follows:
	\begin{equation}
	\label{eqn:softmax}
	\begin{bmatrix}
	P_{N_b} (y = 1 \vert x^b ) \\
	\vdots \\
	P_{N_b} (y = j \vert x^b ) 
	\end{bmatrix}
	=
	\frac{1}{\sum_{p=1}^j e^{w_b^{(p)}N_b'(x)}}
	\begin{bmatrix}
	e^{w_b^{(1)}N_b'(x)} \\
	\vdots \\
	e^{w_b^{(j)}N_b'(x)} 
	\end{bmatrix},
	\end{equation} 
	where, $w_b$ is the weight matrix of the last softmax layer with $w_b^{(p)}$ representing the weight vector that produces the output of the class $p$ and $N_b'(x)$ is the output of the layer in $N_b$, immediately preceding the softmax layer. 
	Once this network is trained, $\cS_b$ broadcasts these models.
	
	At $\cS_i$, a new discriminative model $N_i$ is initialized with the parameters of $N_b$. 
	$N_b$ is trained (and has the ability) to only make predictions on the label space of $\cD_b$, i.e. $\{1 \hdots j\}$. 
	The incremental learner model $N_i$ therefore, cannot be initialized with the same weights in the softmax layer of $N_b$ alone. 
	Along with the weights for the first $j$ classes, $N_i$ should also be initialized with random parameters as necessary to allow for the prediction on a combined incremental label space of $\{1 \hdots c\}$.  
	We can simply do the following assignment to get the desired arrangement:
	\begin{equation}
	\label{eqn:initialize}
	w_i^{(p)} = \begin{cases}
	w_b^{(p)}, 			  & \text{if}  \ \ p \in \{1 \hdots j\} \\
	\mathcal{N}(0,1),  & \text{if} \ \ p \in \{j+1 \hdots c\}
	\end{cases} .
	\end{equation}
	Equation~\ref{eqn:initialize} describes a simple strategy where the weight vectors are carried over to the first $j$ classes and random weight vectors are assigned to the rest of the $c-j$ classes. 
	In figure~\ref{fig:proposed-architecture}, the gray weights in $N_i$ represent those that are copied and the red weights represent the newly initialized weights.
	
	We now have at $\cS_i$, a network that will generate samples from the distribution of $P(x^b)$ and an initialized network $N_i$ whose layers are setup with the weights from $N_b$. 
	To train this network on $\cD_i$, if we simply ignore $G_b$ and train the network with samples $(x^i,y^i) \in \cD_i$, we will run into the catastrophic forgetting problem as discussed in figure~\ref{fig:forgetting}. 
	To avoid this, we can use samples queried from $G_b$ (such samples are notationally represented as $G_b(z)$ to indicate sampling using a random vector $z$) and use these samples to avoid forgetting. 
	However we do not have targets for these samples to estimate an error with. 
	Phantom sampling will help us to acquire targets.
	
	\begin{definition}
		A phantom sampler is a process of the following form:
		\begin{equation}
		\mathcal{P}: (z, T, N_b, G_b) \rightarrow \{G_b(z), P_{N_b}(y \vert G_b(z),T)\}.
		\end{equation}
	\end{definition}
	where, $y \in \{0 \hdots j\}$ and $T$ is a temperature parameter which will be described below.
	Using $N_b$ and $G_b$, we can use this sampling process to generate sets of sample-target pairs that simulate samples from the dataset $\cD_b$. 
	Simply using $P_{N_b}(y^b \vert x^b)$ is not possible as we do not have access to $x^b$ at $\cS_i$, and $\cS_i$ is not allowed to communicate with $\cS_b$ regarding the data due to the data membrane condition described in property~\ref{prop:membrane}.
	We can however replace $x^b$ with $G_b(z)$ and use the generated samples to produce targets from this network for the generated samples itself. 
	This is justifiable since $G_b(z)$ is learnt to hallucinate samples from $P(x^b)$. 
	However, given that we only use a simple GAN and that the samples are expected to be noisy, we might get corrupted and untrustworthy targets. 
	GANs have not advanced sufficiently to a degree where perfect sampling is possible at the image level, at the moment of writing this article.
	As GAN technology improves, much better sampling could be achieved using this process. 
	
	Given that GANs (and any other similar generative models) are imperfect, often samples can have properties that are blended from two or more classes. 
	In these cases, the targets generated from $N_b$ might also be too high for  only one of these classes, which is not optimal. 
	To avoid this problem, we use a replacement for the softmax layer of $N_b$ with a new temperature-raised softmax layer,
	\begin{equation}
	\label{eqn:temp_softmax}
	\begin{bmatrix}
	P_{N_b} (y = 1 \vert x^b, T ) \\
	\vdots \\
	P_{N_b} (y = j \vert x^b, T  ) 
	\end{bmatrix}
	=
	\frac{1}{\sum_{p=1}^j e^{\frac{w_b^{(p)}N_b'(x)}{T}}}
	\begin{bmatrix}
	e^{\frac{w_b^{(1)}N_b'(x)}{T}} \\
	\vdots \\
	e^{\frac{w_b^{(j)}N_b'(x)}{T}} 
	\end{bmatrix}.
	\end{equation} 
	This temperature-raised softmax for $T>1$ ($T=1$ is simply the softmax described in equation~\ref{eqn:softmax}) provides a softer target which is smoother across the labels. 
	It reduces the probability of the most probable label and provides rewards for the second and third most probable labels also, by equalizing the distribution. 
	Soft targets such as the one described and their use in producing ambiguous targets exemplifying the relationships between classes were proposed in~\cite{hinton2015distilling}. 
	In this context, the use of soft targets for $G_b(z)$ helps us get appropriate labels for the samples that may be poorly generated. 
	For instance, a generated sample could be in between classes $8$ and $0$. 
	The soft target for this will not be a strict $8$ or a strict $0$, but a smoother probability distribution over the two (all the) classes. 
	
	While learning $N_i$, with a batch of samples from $D_i$, we may simply use a negative log-likelihood with the softmax layer for the labels. 
	To be able to back-propagate samples from phantom sampling, we require a temperature softmax layer at $N_i$ as well. 
	For this, we simply create a temperature softmax layer that share the weights $w_i$, of the softmax layer of $N_i$, just as we did for $N_b$. 
	This implies that $N_i$ will have $c - j+1$ additional units for which we would not have targets as phantom sampling will only provide us with targets for the first $j$ classes. 
	Given that the samples themselves are hallucinated from $G_b(z)$, the optimal targets to assign for the output units $[j+1 \hdots c]$ of the temperature softmax layer are zero. 
	Equivalently, we could simply avoid sharing the extra weights. 
	Therefore along with the phantom sample's targets, we concatenate a zero vector of length $[j+1 \hdots c]$. 
	This way, we could simply back-propagate the errors for the phantom samples also. 
	The error for data from $\cD_i$ is,
	\begin{equation}
	e(w_i,x^i \in \cD_i) = \mathcal{L} ( y^i, \argmax_{y} P_{N_i}(y \vert x^i) ),
	\end{equation}
	where, $\mathcal{L}$ represents an error function. 
	The error for phantom samples is,
	\begin{equation}
	e(w_i,G_b(z)) = \mathcal{L} ( P_{N_b}(y \vert G_b(z), T), P_{N_i}(y \vert G_b(z), T)).
	\end{equation}
	Typically, we use a categorical-cross-entropy for learning labels and a root mean-squared error for learning soft-targets.
	
	While both samples from $\cD_i$ and from the phantom sampler are fed-forward through the same network, the weights are updated for two different errors. If the samples come from the phantom sampler, we estimate the error from the temperature softmax layer and if the samples come from $\cD_i$, we estimate the errors from the softmax layer. 
	For every $k$ iterations of $\cD_b$, we train with $1$ iteration of phantom samples $G(z)$. 
	$k$ is decided based on the number of classes that are in each set $\cD_b$ and $\cD_i$. 
	
	Thus far we have  assumed a certain chronology of events where $\cS_i$ begins training only after $\cS_b$ is finished training. 
	We could generalize this strategy of using phantom sampling when $\cS_i$ is already, partially trained by the time $\cS_b$ finishes and triggers the incremental learning. 
	In this case, we will not be able to re-initialize the network $N_i$ with new weights, but as long as we have phantom samples, we can use a technique similar to mentor nets or fitnets, using embeded losses between $N_b$ and $N_i$ and transfer knowledge about $D_b$ to $N_i$ ~\cite{romero2014fitnets}~\cite{venkatesan2016diving}. 
	This strategy could also be extended to more than one increment of data in a straight-forward manner. 
	Using the same phantom sampling technique we could continue training the GAN to update it with the distributions of the new classes.
	Once trained, we can pass on this GAN and the newly trained net $N_i$ to the next incremental site.

	\section{Related Work}
	\label{sec:lit}
	\begin{figure*}[!ht]
		\begin{center}
			\includegraphics[width=0.99\linewidth]{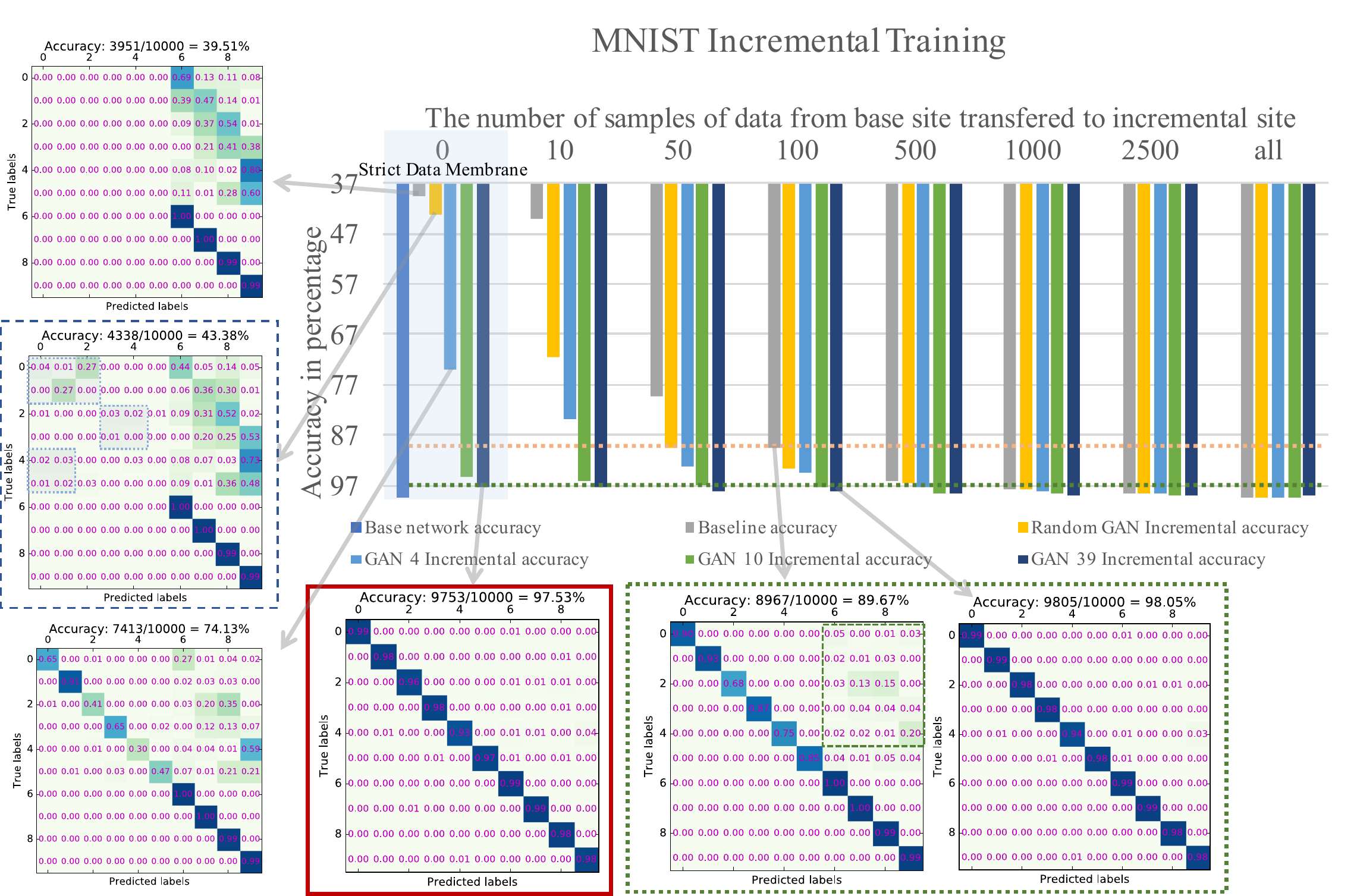}
		\end{center}
		
		\caption{Results for the MNIST dataset. Base network is $N_b$, baseline is $N_i$ without phantom sampling. GAN $q$ is phantom sampling with GAN trained for $q$ epochs.}
		
		\label{fig:mnist-results}
	\end{figure*}
	\noindent\textbf{Catastrophic Forgetting}:
	Early works by McCloskey, French and Robins outlines this issue \cite{mccloskey1989catastrophic, french1993catastrophic, robins1995catastrophic}. 
	In recent years, this problem has been tackled using special activation functions and dropout regularization. 
	Srivastava et al. demonstrated that the choice of activation function affects catastrophic forgetting and introduced the \emph{Hard Winner Take All} (HWTA) activation \cite{srivastava2013compete}. 
	Goodfellow et al. argued that increased dropout works better at minimizing catastrophic forgetting compared to activation functions \cite{goodfellow2013empirical}. 
	All these studies were made in regards to unavailability of data for particular classes, rather than in terms of incremental learning.
	
	We find that most previous works in incremental learning, relaxes or violates the rigorous constraints that we have proposed for an incremental learner. While this may satisfy certain case studies, pertaining to each article, we find no work that has addressed our definition sufficiently. In this section, we organize our survey of existing literature in terms of the conditions they violate.
	
	\noindent\textbf{Relaxing the data membrane}:
	The following approaches relax property (\ref{prop:membrane}) to varying degrees. 
	Mensink et al. develop a metric learning method to estimate the similarity (distance) between test samples and the \emph{nearest class mean} (NCM) \cite{mensink2012metric, mensink2013distance}. 
	The class mean vectors represent the centers of data samples belonging to different classes. 
	The learned model is a collection class center vectors and a metric for distance measurement that is determined using the training data. 
	The NCM approach has also been successfully applied to random forest based models for incremental learning in \cite{ristin2014incremental}. 
	The nodes and leaves of the trees in the NCM forest are dynamically grown and updated when trained with data from new classes. 
	A tree of deep convolutional networks (DCNN) for incremental learning was proposed by Xiao et al. \cite{xiao2014error}. 
	The leaves of this tree are CNNs with a subset of class outputs and the nodes of the tree are CNNs which split the classes. 
	With the input of new data and classes, the DCNN grows hierarchically to accommodate the new classes. 
	The clustering of classes, branching and tree growth is guided by an error-driven preview process and their results indicate that the incremental learning strategy performs better than a network trained from scratch. 
	
	The Learn++ is an ensemble based approach for incremental learning \cite{polikar2001learn} \cite{muhlbaier2009learn}. 
	Based on the Adaboost, the algorithm weights the samples to achieve incremental learning. 
	The procedure, however requires every data batch to have examples from all the previously seen classes. 
	In \cite{kuzborskij2013n}, Kuzborskij et al. develop a least squares SVM approach to incrementally update a N-category classifier to recognize N+1 classes. 
	The results indicate that the model performs well only when the N+1 classifier model is also trained with some data samples from the previous N classes.
	
	iCaRL is an incremental representation based learning method by Rebuffi et al. \cite{rebuffi2016icarl}. 
	It progressively learns to recognize classes from a stream of labeled data with a limited budget for storing exemplars. 
	The iCaRL classification is based on the nearest-mean-of-exemplars. 
	The number of exemplars for each class is determined by a budget and the best representation for the exemplars is updated with existing exemplars and newly input data. 
	The exemplars are chosen based on a herding mechanism that creates a representative set of samples based on a distribution \cite{welling2009herding}. 
	This method while being very successful, violates the membrane property by transferring well-chosen exemplar samples.
	In our results section we address this idea by demonstrating that significant amount of (randomly chosen) samples are required to out-perform our strategy, which violates the \emph{budget} criteria of the iCaRL methods. 
	
	\noindent\textbf{Relaxing data agnosticism}: 
	Incremental learning procedures that draw inference regarding previously trained data based on current batch of training data, can be viewed as violating this constraint. 
	Li et al. use the base classifier $N_b$ to estimate the conditional probabilities $P(\hat{y}|x)$ for $x:(x,y)\in\cD_i$. 
	When training $N_i$ with $\cD_i$, they use these conditional probabilities to guide the output probabilities for classes $y\in[1,\ldots, j]$ \cite{li2016learning}. 
	In essence, the procedure assumes that if $N_i$ is trained in such a manner that $P(\hat{y}|x)$ for $x:(x,y)\in\cD_i$ is the same for both classifier $N_b$ and $N_i$, this ensures that $P(\hat{y}|x)$ for $x:(x,y)\in\cD_b$ will also be the same. 
	This is a strong assumption relating $\cD_b$ and $\cD_i$ violating agnosticism. 
	The authors Furlanello et al. develop a closely related procedure to in \cite{furlanello2016active}. 
	They train neural networks for the incremental classifier $N_i$ by making sure the conditional probabilities $P(\hat{y}|x)$ for $x:(x,y)\in\cD_i$ is the same for both $N_b$ and $N_i$. 
	The only difference compared to \cite{li2016learning} is in the regularization of network parameters using weight decay and the network initialization. 
	In another procedure based on the same principles, Jung et al. constrain the feature representations for $\cD_i$ to be similar to the feature representations for $\cD_b$ \cite{jung2016less}. 
	
	Other models assume that the parameters of the classifiers $w_b$ for $N_b$ and $w_i$ for $N_i$ are related. 
	Kirkpatrick et al. model the probability $P(w_b|\cD_b)$ and get an estimate for the important parameters in $w_b$ \cite{kirkpatrick2016overcoming}. 
	When training $N_i$ initialized with parameters $w_b$, they make sure not to offset the important parameters in $w_b$. 
	This compromises the training of $N_i$ under the assumption that important parameters in $w_b$ for $\cD_b$ are not important for $\cD_i$.  
	
	Closely related to the previous idea is \emph{pseudo-rehearsal} proposed by Robins in 1995~\cite{robins1995catastrophic}. 
	Neuro-biological underpinnings of this work was also studied by French et. al,~\cite{french1997pseudo}. 
	This method is a special case of ours if, the GAN was untrained and produces random samples.
	In other words, they used $N_b$ to produce targets for random samples $G_b(z) = z \rightarrow \mathcal{N}(0,1)$, instead of using a generative model, similar to phantom sampling. 
	This might partly be due to the fact that sophisticated generative models were not available at the time. 
	This article also does not use soft targets such as those that we use because, for samples that are generated randomly, $T=1$ is a better target. This article does not violate any of the properties that we required for our uncompromising incremental learner.
	
	\section{Experiments and Results}
	\label{sec:res}
	
	To demonstrate the effectiveness of our strategy we conduct thorough experiments using three benchmark datasets: MNIST dataset of handwritten character recognition, Street view housing numbers (SVHN) dataset and the CIFAR10 10-class visual object categorization dataset~\cite{mnist, netzer2011reading, krizhevsky2009learning}. 
	In all our experiments\footnote{Our implementations are in theano and our code is available at \url{https://github.com/ragavvenkatesan/Incremental-GAN}.} we train the $\cS_b$'s GAN, $G_b$ and base networks $N_b$ independently. 
	The network parameters of all these models are written to drive, which simulates broadcasting the networks.
	Once trained, the datasets that are used to train and test these methods are deleted, simulating the data membrane and the processes are killed. 
	
	We then begin $\cS_i$ as an independent process in keeping with site independence. 
	This uses a new dataset which is setup in accordance with property~\ref{prop:membrane}. 
	Networks $G_b$ and $N_b$'s parameters are loaded but only in their feed-forward operations. 
	Two identical copies of networks $N_i^\sigma$ and $N_i^T$ that share weights are built. 
	These are initialized with the parameters of $N_b$,  $N_i^\sigma$with without temperate and $N_i^T$ with temperature softmax layers. 
	By virtue of the way they are setup, updating the weights on one, updates both the networks. 
	We feed forward $k$ mini batches of data from $\cD_i$ through the column that connects to the softmax layer and use the error generated here to update the weights for each mini batch. 
	For every $k$ updates of weights from the data, we update one mini batch of phantom samples from $(G_b(z), P_{N_b}(y \vert G_b(z),T))$. 
	This is run until early termination or until a pre-determined number of epochs. 
	Since we save the parameters of $G_b$ after every epoch, we can load the corresponding GAN for our experiments. 
	We use the same learning rate schedules, optimizers and momentums across all the architectures.
	We fix our temperature values using a simple grid search.
	We conducted several experiments using the above protocol to demonstrate the effectiveness of our strategy.
	The following sections discuss these experiments. 
	
	\begin{figure}[t]
		\begin{center}
			\includegraphics[width=0.99\linewidth]{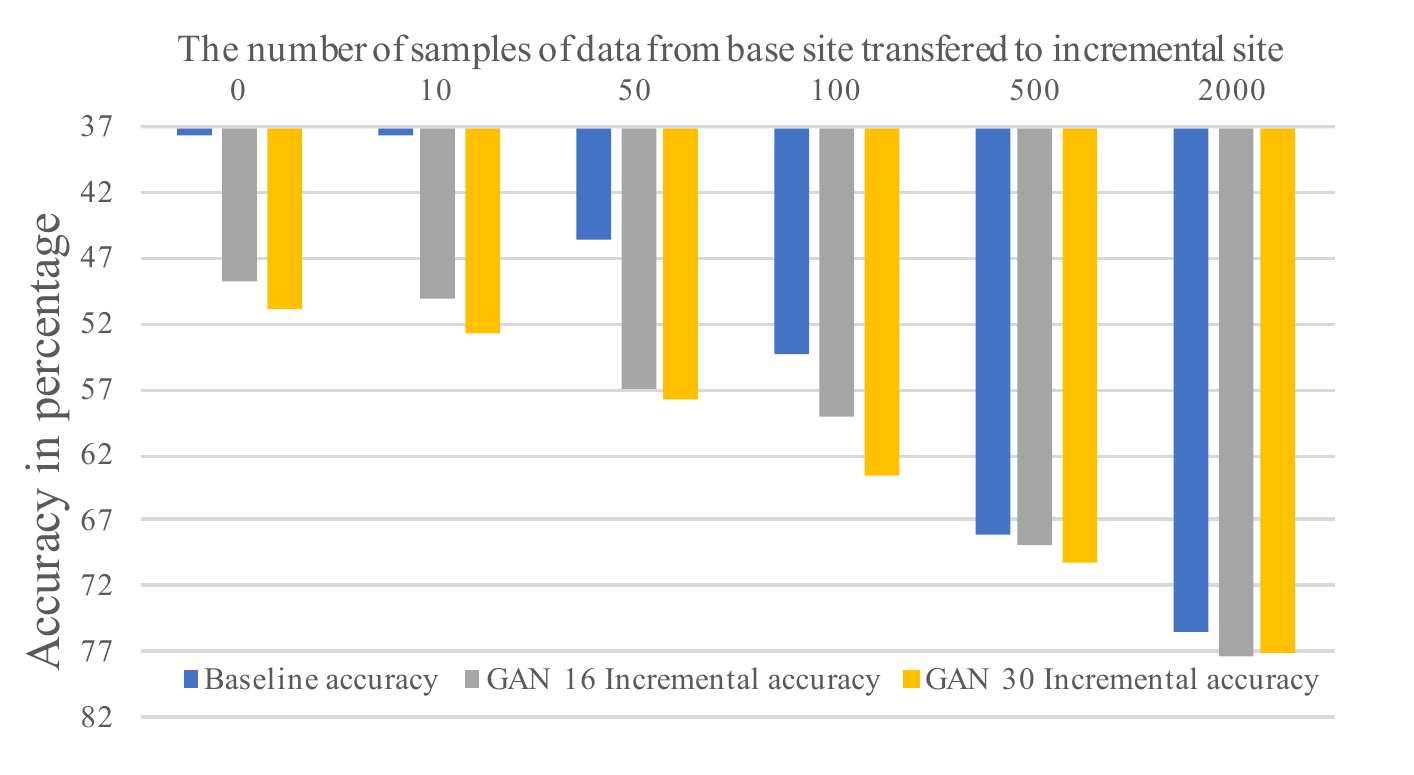}
		\end{center}
		
		\caption{Results for the CIFAR10 dataset. Notation, similar to that of figure~\ref{fig:mnist-results}.}
		
		\label{fig:cifar-results}
	\end{figure}
	
	\begin{figure}[t]
		\begin{center}
			\includegraphics[width=0.99\linewidth]{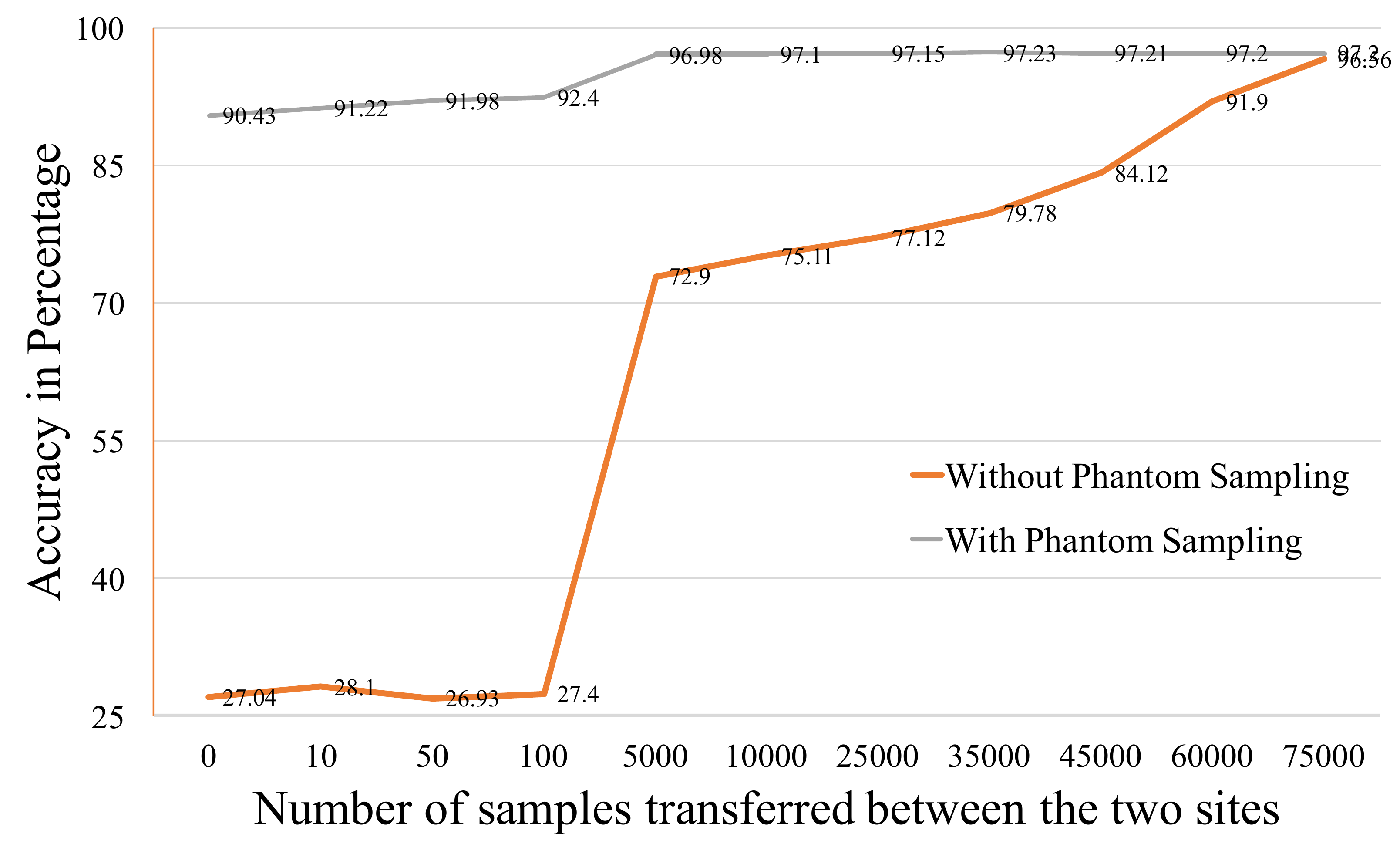}
		\end{center}
		
		\caption{Results for the SVHN dataset using a well-trained GAN.}
		
		\label{fig:svhn-results}
	\end{figure}
	
	\subsection{Single dataset experiments}
	
	\noindent\textbf{MNIST:} For the MNIST dataset,  we used a GAN $G_b$ that samples $10$ image generations from a uniform $0$-mean Gaussian. 
	The generator part of the network has three fully-connected layers of $1200$, $1200$ and $784$ neurons with ReLU activations for the first two and tanh activation for the last layers, respectively~\cite{nair2010rectified}. 
	The discriminator part of $G_b$ has two layers of $240$ maxout-by-$5$ neurons~\cite{goodfellow2013maxout}. 
	This architecture that mimics the one used by Goodfellow et. al, closely~\cite{goodfellow2014generative}. 
	All our discriminator networks across both sites $\cS_b$ and $\cS_i$ are the same architecture which for the MNIST dataset is, two convolutional layers of $20$ and $50$ neurons each with filter sizes of $5 \times 5$ and $3 \times 3$ respectively, with max pooling by $2$ on both layers. 
	These are followed by two full-connected layers of $800$ neurons each. 
	All the layers in the discriminators are trained with batch normalization and weight decay with the fully-connected layers trained with a dropout of $0.5$~\cite{srivastava2014dropout, ioffe2015batch}. 
	
	Results of the MNIST dataset are discussed in figure~\ref{fig:mnist-results}.
	The bar graph is divided into many factions $p = [0, 10, \dots \text{all}]$, each representing the performance having $p$ samples per class transmitted between $\cS_b$ to $\cS_i$. 
	Within each faction are five bars, except $p=0$ that has six bars.
	The first bar at $p=0$ represents the state-of-the-art accuracy with the (base) network trained on the entire dataset ($\cD_b \cup \cD_i$, for the given hypothesis.
	This is the upper-bound on the accuracies, given the architecture.
	The first bar on the left (second for $p=0$) represents the accuracy of a \emph{baseline} network that is learnt without using our strategy. 
	A baseline network does not use a phantom sampler and is therefore prone to catastrophic forgetting.
	The other four bars represent the performance of networks learnt using our strategy.
	From left to right, the $G_b$ for each network is trained for $e = [0, 4, 10, 39]$ epochs, respectively. 
	Confusion matrices are shown wherever appropriate. 
	
	The central result of this experiment is the block of accuracies highlighted within the blue-shaded box ($p=0$), which show the performances while maintaining a strict data membrane.
	The confusion matrix in the top-left corner shows the performance of the base network with $p=0$, which is similar to (c) from figure~\ref{fig:forgetting}, demonstrating catastrophic forgetting. 
	The next confusion matrix that is marked with blue dashed line depicts the accuracy of $N_i$ with $G_b$ producing random noise.
	This setup is the same as in the work by Robins~\cite{robins1995catastrophic}.  
	It can be observed that even when using a phantom sampler that samples pure noise, we achieve a noticeable boost in recognition performance, significantly limiting catastrophic forgetting.
	The confusion matrix in the bottom-left corner is the performance using $G_b$ trained for only $4$ epochs.
	This shows that even with a poorly trained GAN, we achieve a marked increase in performance.
	The best result of this faction is the confusion matrix highlighted in the red square.
	This is the result of a network learnt with phantom sampling with a GAN $G_b$ that is trained closest to convergence at $39$ epochs. 
	It can be clearly noticed that the phantom sampling strategy helps in avoiding catastrophic forgetting, going so far as to achieve nearly state-of-the-art base accuracy.   
	
	The rest of the factions in this experiment make a strong case against the relaxation of the data membrane.  
	Consider, for instance, the pair of confusion matrices at the bottom right, highlighted within the green dotted lines.
	These represent the performance of \emph{baseline} and $e=39$ networks, when $p=100$ samples per-class were transmitted through the membrane.
	A \emph{baseline} network that was trained carefully without overfitting produced an accuracy of $89.67\%$ and still retained a lot of confusion (shown in green dashed lines within the confusion matrix). 
	The network trained with phantom sampling significantly outperforms this. 
	In fact (refer the orange dotted line among the bars), this relaxation is outperformed by a phantom sampling trained network even with a poorly trained GAN (with just $10$ epochs) while adhering to a strict data membrane ($p=0$). 
	It is only when $p = 1000$ samples per-class (which is $20\%$) of the data are being transferred, does the baseline even match the phantom sampling network with $p=0$ (as demonstrated by the blue dotted line among the bars). 
	All these results conclusively demonstrate the significance of phantom sampling and demonstrate the nonnecessity of the relaxation of the data membrane. 
	An uncompromising incremental learner was thereby achieved using our strategy.
	
	\noindent\textbf{SVHN and CIFAR 10:} For both these datasets we used a generator model that generates images from $64$ Gaussian random variables. 
	The number of neurons in subsequent fully-connected layers are $1200$ and $5408$ respectively. 
	This is followed by two fractionally-strided or transposed convolution layers with filter sizes $3 \times 3$ and $5 \times 5$ respectively. 
	Apart from the last layer that generates the $32 \times 32$ image, every layer has a ReLU activation. 
	The last layer uses a tanh activation. 
	Our discriminator networks including the discriminator part of the GANs have six convolutional layers with neurons $20, 50, 50, 100, 100$ and $250$ respectively. 
	Except the first layer, which has a filter size of $5 \times 5$, every layer has filter sizes of $3 \times 3$.
	Every third layer maxpools by $2$. 
	These are followed by two fully-connected layers of $1024$ nodes each. 
	All activations are ReLU.
	
	Results of the CIFAR 10 dataset are shown in figure ~\ref{fig:cifar-results} and that of SVHN are shown in figure~\ref{fig:svhn-results}.
	CIFAR 10 and SVHN contain three channel full-color images that are sophisticated.
	GANs, as originally proposed by Goodfellow et. al, fail to generate reasonably good looking samples for these datasets~\cite{goodfellow2014generative}.
	Since we used the same models, the results shown here could be improved significantly with the invention (or adoption) of better generative models.
	
	We can observe from figures~\ref{fig:cifar-results} and~\ref{fig:svhn-results}, that they follow patterns similar to the MNIST results in figure~\ref{fig:mnist-results}.
	The CIFAR-10 results clearly demonstrate that only after about $20\%$ of data is transmitted, do the performance come close to matching the phantom sampler approach. 
	In the SVHN results shown in figure~\ref{fig:svhn-results}, we can observe the marked difference in performance with only few samples being transmitted.
	Because SVHN is a large dataset in number of samples, the GANs were able to generate sufficiently good images that lead to superior performance.
	This result shows us the advantage our strategy has when working with big datasets.
	Firstly, having a big dataset imposes additional penalties for requiring to transmit data and therefore should be avoided. 
	Secondly, having more number of samples implies that a simple GAN could generate potentially good looking images, helping us maintain consistent performance throughout.

	\subsection{Cross-domain increments}
	
	\begin{figure}[t]
		\begin{center}
			\includegraphics[width=0.99\linewidth]{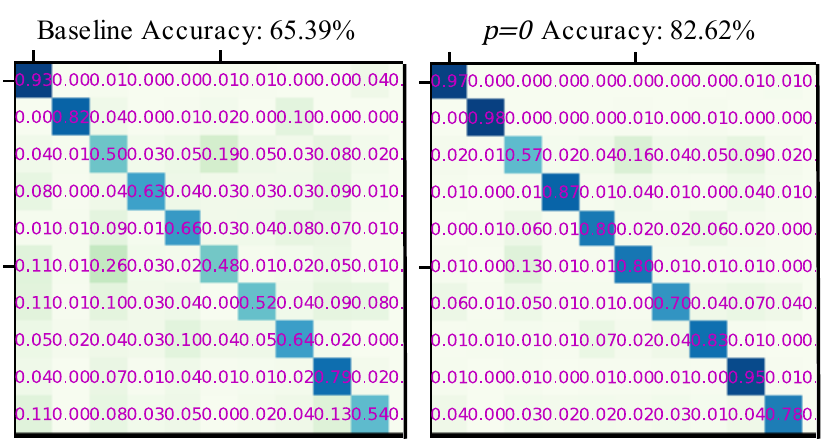}
		\end{center}
		
		\caption{Results for MNIST-rotated trained at $\cS_b$ and incremented with new data from the MNIST original dataset at $\cS_i$. The  class labels for both these datasets is $[0, \dots 9]$. The confusion matrix on the left is for the baseline network and the one on the right is for our strategy with $p=0$.}
		
		\label{fig:mnist-cross}
	\end{figure}
	
	\begin{figure}[t]
		\begin{center}
			\includegraphics[width=0.99\linewidth]{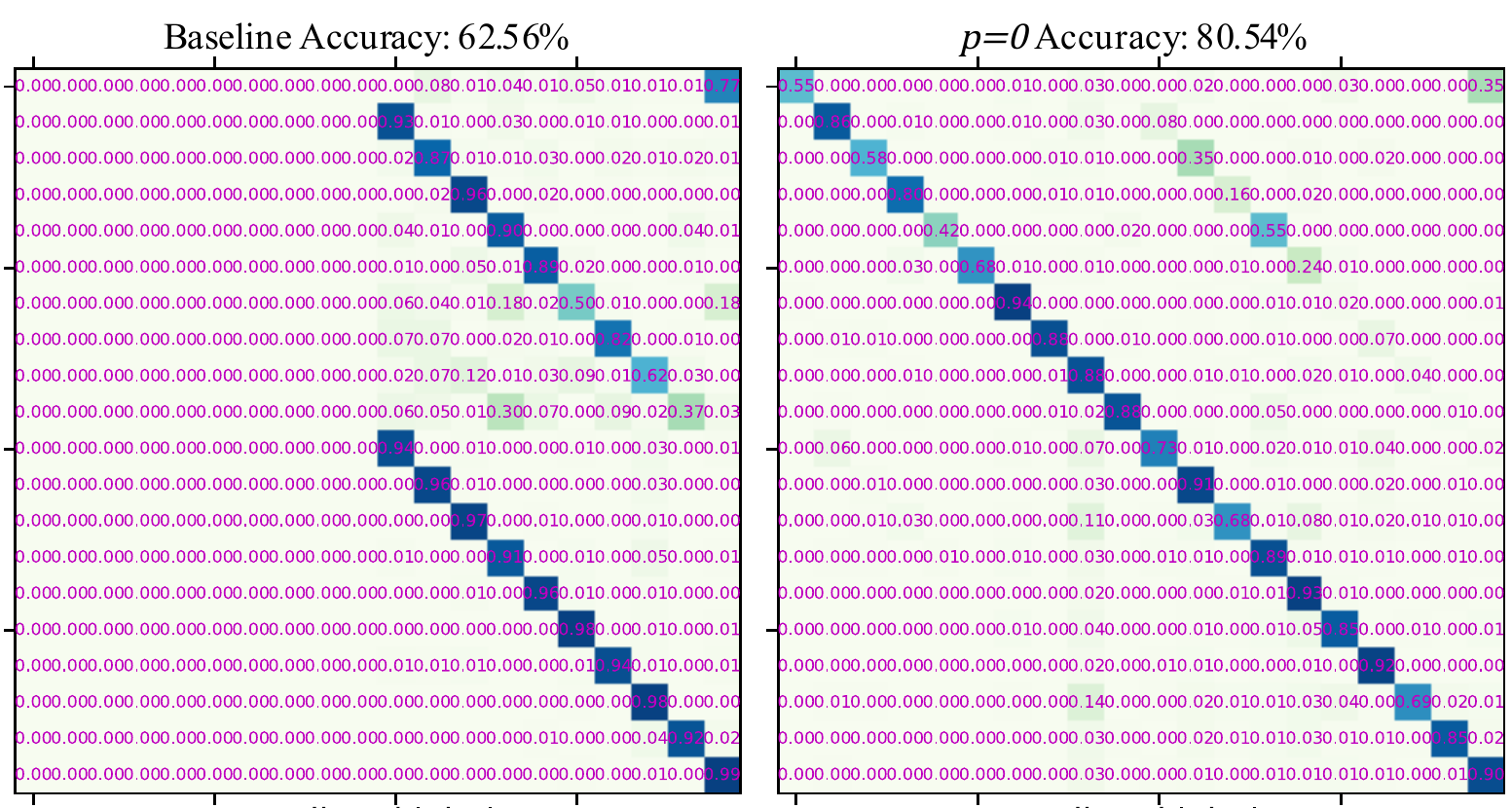}
		\end{center}
		
		\caption{Results for MNIST trained at $\cS_b$ and incremented with new data from the SVHN dataset at $\cS_i$. The SVHN classes are considered as novel classes in this experiment, therefore we have twenty classes. The confusion matrix on the left is for the baseline network and the one on the right is for our strategy with $p=0$.}
		
		\label{fig:mnist-svhn}
	\end{figure}
	
	It could be argued that performing incremental learning within the same dataset has some advantages in terms of the domain of the datasets being similar.
	The similarity in domains could imply that the datasets are general and therefore, the base network already has some features of the incremental dataset encoded in it~\cite{venkatesan2016neural}.  
	In this section we demonstrate two special cross-domain cases.
	In the first case, the incremental data $\cD_i$, while sampled from a new domain, has the same label space as $\cD_b$ .
	In the second case, $\cD_i$ has new classes that are not seen in $\cD_b$.

	\begin{figure*}[t]
		\begin{center}
			\includegraphics[width=0.99\linewidth]{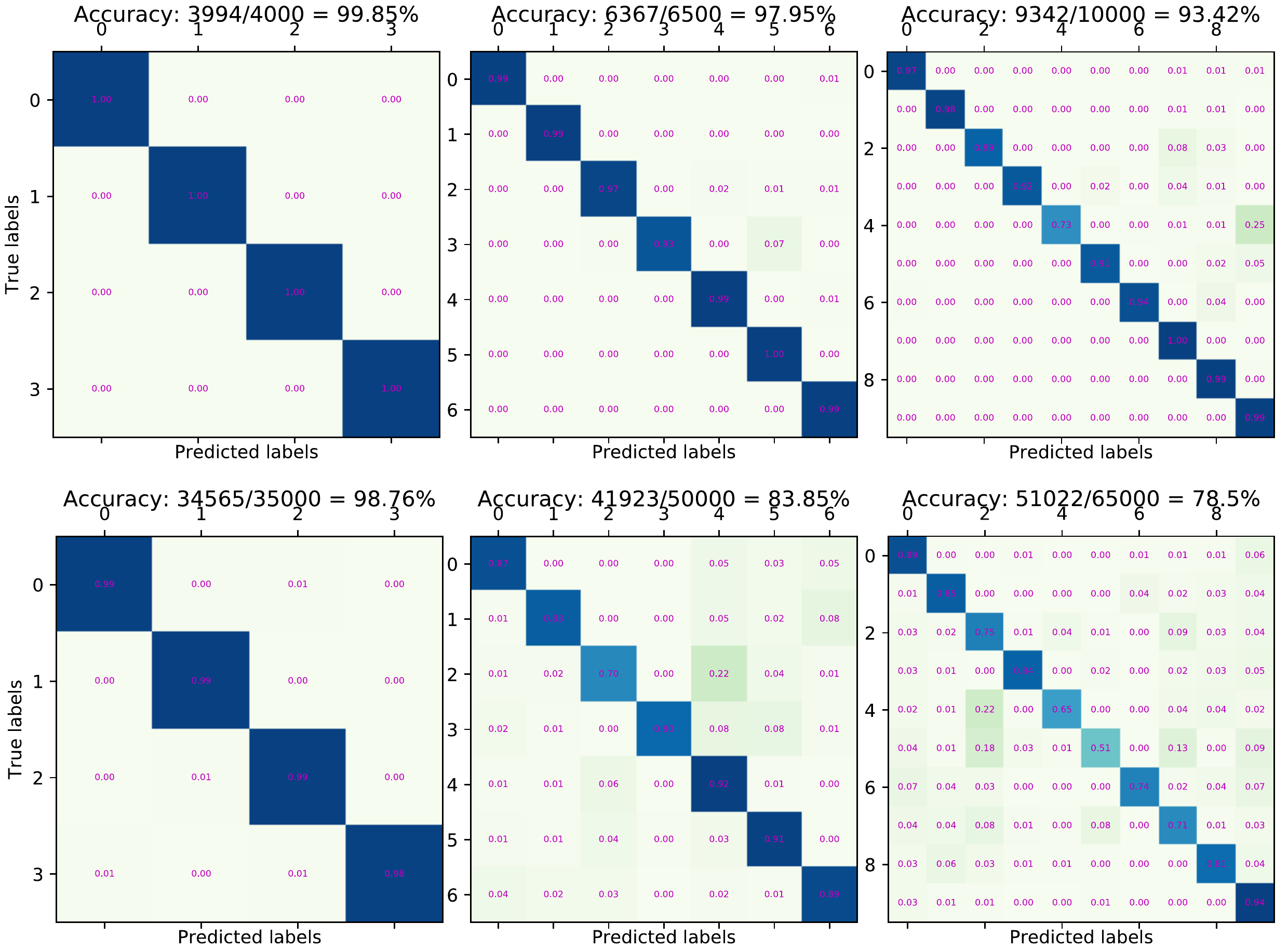}		
		\end{center}
		
		\caption{Results for the bounded-continual learning experiments. There are two steps of increment. Each increment has its own GAN. The top row is MNIST and the bottom row is SVHN. In each row, the image on the left is the confusion of the base net $N_0$ with classes $[0,1,2,3]$. The center image is the confusion for the first increment with training data in classes $[4,5,6]$ and testing data in classes $[0, \dots 6]$. The confusion on the right is the final increment with training data from classes $[7,8,9]$ and testing data from the classes $[0, \dots 9]$.}
		
		\label{fig:continual}
	\end{figure*}
	\noindent\textbf{Case 1:}
	In this experiment, our base dataset $\cD_b$ is the MNIST-rotated dataset developed by Larochelle et al.~\cite{larochelle2007empirical}.
	This is used to learn $G_b$ and $N_b$.
	This is a dataset that is the same as the MNIST dataset, but the samples are randomly rotated.
	The incremental data comes from the MNIST dataset .
	The incremental data and the base dataset has the same label space.
	The domain of incremental dataset $\cD_i$ (MNIST) can be considered as a special subset of the domain of $\cD_b$ (MNIST-rotated). 
	Therefore, this setup is ripe for a scenario where the incremental site forgets the expanse of the domain of the base site. 
	The network architecture remains the same as for the MNIST experiments.
	The results for this experiment are shown in figure~\ref{fig:mnist-cross}.
	It can be clearly noted that there is about $20\%$ difference in performance using our strategy.
	
	\noindent\textbf{Case 2:}
	In this experiment, our base dataset $\cD_b$ is the MNIST dataset and it is used to learn $G_b$ and $N_b$.
	The incremental dataset $\cD_i$ is SVHN. 
	The classes of SVHN are labelled $10 - 19$ at $\cS_i$ and the labels of MNIST are maintained as $0-9$.
	This is essentially incrementing on a new task from a disjoint domain.
	The results of this experiment are shown in figure~\ref{fig:mnist-svhn}.
	It can be clearly noted that there is about $20\%$ increase in performance using our strategy.
	
	\section{Extension to bounded-continual learning}
	So far we have defined and studied incremental learning. 
	Incremental learning consists of a single increment. 
	In this section, we extend this idea to bounded-continual learning.
	Continual learning is incremental learning with multiple increments. 
	Bounded-continual learning is a special case of continual learning, where the number of increments is limited. 
	Life-long learning for instance, is an example of unbounded-continual learning. 
	
	The proposed strategy can be trivially modified to work for multiple increments.
	Consider there are $s$ sites.
	Consider also that we have one base network $N_b^i$, with $i$ indicating its state after the increment $i$.
	We learn for every increment $i$, a new GAN $G_i$. 
	We use the set of GANs $\{G_0, \dots G_{i-1}\}$ to create $i$ phantom samplers, one for each increment. 
	
	Continual learning can be implemented in the following manner.
	At the beginning, we construct a base network $N_b^0$. 
	Once $N_b^0$ is trained with $\cD_0$, we create a copy ($P^0$) of $N_b^0$ for phantom labelling.
	The samples generated by $G_0$ are fed through $P^0$, to get phantom samples for the increment $i = 0$.
	This phantom sampler will be used when learning the increment $i=1$ .
	
	On receiving the data increment $\cD_i$, we have $i$ GANs $G_0, \dots G_{i-1}$.
	We can create an updated copy of the phantom sampler $P^{i-1}$, by making a copy of $N_b^{i-1}$. 
	We create a phantom sampler, where $P^{i-1}$ samples from all the GANs uniformly and hallucinates the labels. 
	We update $N_b^{i-1}$ to $N_b^{i+1}$, by training it on $\cD_i$ along with this new phantom sampler $P^{i-1}$.
	
	This approach of bounded-continual learning is apt in cases where the data at each increment is large enough to warrant training a GAN. 
	While, this approach works well for bounded-continual learning systems, it is not scalable to lifelong learning. 
	This is because unbounded-continual learning could result in an infinite number of GANs. 
	Seff et al, recently proposed an idea to update the same GAN for a large number of increments~\cite{seff2017continual}.
	Such a GAN could generate data from the combined distributions of all increments it has seen. 
	While this still works only on a bounded number of increments, this is a step towards unbounded-continual learning.
	If we employ this idea in our system, we could eliminate the need for having multiple GANs and extend our strategy trivially to life-long learning as well. 
	This idea is still in its infancy and is not fully mature yet.
	Although we have drawn a road map, we await further development of this idea to incorporate it fully into our strategy.
	
	\label{sec:continual}
	\subsection{Experiments and results}
	We use GANs and classifier architectures which are the same as defined for MNIST and SVHN in the previous section, respectively.
	We demonstrate continual learning on both datasets by performing two increments.
	The base dataset contains the classes $[0,1,2,3]$, the first increment contains classes $[4,5,6]$ and the last increment contains $[7,8,9$]. 
	Figure~\ref{fig:continual} shows the results for continual learning for both datasets.
	It can be easily noticed that we can achieve close to state-of-the-art accuracy even while performing continual learning. 
	A note of prominence is that even at the end of the third increment, there is little confusion remaining from the first increment. 
	This demonstrates strong support for our strategy even when extending to continual learning. 
	
	\section{Conclusions}
	\label{sec:conclusion}
	
	In this paper, we redefined the problem of incremental learning, in its most rigorous form so that it can be a more realistic model for important real-world applications. 
	Using a novel sampling procedure involving generative models and the distillation technique, we implemented a strategy to hallucinate samples with appropriate targets using models that were previously trained and broadcast. 
	Without having access to historic data, we demonstrated that we could still implement an uncompromising incremental learning system without relaxing any of the constraints of our definitions. We show strong and conclusive results on three benchmark datasets in support of our strategy. 
	We further demonstrate the effectiveness of our strategy under challenging conditions, such as cross-domain increments, incrementing label space and bounded-continual learning.

{\small
	\bibliographystyle{ieee}
	\bibliography{egbib}
}	
\end{document}